# Human–AI Collaborative Inductive Thematic Analysis: How AI Guides Analysis and Researchers Reclaim Interpretive Authority

**Matthew Nyaaba**[†1,2], **Min SungEun**[†3], **Mary Abiswin Apam**[†4], **Kwame Owoahene Acheampong**[5], **Emmanuel Dwamena**[6], **Xiaoming Zhai**[1, 7]

[1]AI4STEM Education Center, University of Georgia, Athens, GA, USA
[2]Department of Educational Theory and Practice, University of Georgia, Athens, GA, USA
[3]Department of Elementary, Middle Level, Library and Technologies Education, Kutztown University, Kutztown, PA, USA.
[4]Department; Midwifery Department, College of Nursing, Midwifery and Allied Health Sciences, Nalerigu, Ghana.
[5]Department of Lifelong Education, Administration, and Policy, University of Georgia, Athens, GA, USA
[6]Department of Curriculum and Instruction, University of Connecticut, USA
[7]Department of Science, Social Studies and Mathematics, University of Georgia, Athens, GA, USA

Abstract

The growing integration of generative artificial intelligence (GenAI) into qualitative research necessitates closer examination of how such tools shape analytic practice, judgment, and epistemic authority. This study investigates how researchers interact with an Inductive Thematic Analysis GPT (ITA–GPT), a purpose-built AI tool designed to operationalize established inductive thematic analysis procedures through interactive semi-auto prompts aligned with Braun and Clarke's reflexive thematic analysis and Naeem's verbatim coding principles. Guided by a Human–Artificial Intelligence Collaborative Inductive Thematic Analysis (HACITA) framework, grounded in reflexive thematic analysis, human-in-the-loop AI, and distributed cognition, the study focuses on analytic process rather than substantive content. Three experienced qualitative researchers independently conducted ITA–GPT–assisted analyses of interview transcripts from education research in the Ghanaian teacher education context. The ITA–GPT tool guided analysts through familiarization, verbatim (in-vivo) coding, gerund-based descriptive coding, and theme development, while enforcing trace-to-text integrity, transcript coverage checks, and auditability. Data sources included interaction logs, AI-generated tables, researcher revisions, comments, deletions, insertions, and reflexive memos. Findings show that ITA–GPT functioned as a procedural and methodological scaffold, structuring analytic workflow and enhancing transparency. However, researchers consistently exercised epistemic authority through five recurrent analytic actions: modification, deletion, rejection, insertion, and commenting. These actions were essential for correcting AI literalism, restoring contextual and emotional nuance, strengthening audit trails, and aligning interpretations with institutional and professional realities. Verbatim codes were perceived as the most reliable analytic foundation, while AI-generated abstractions required systematic human refinement. The study contributes a theoretically grounded account of how inductive thematic analysis is enacted through human–AI collaboration.

**Keywords:** Human–AI collaboration; Generative artificial intelligence (GenAI); Inductive thematic analysis; Qualitative data analysis; Epistemic authority; Reflexive methodology

## HIGHLIGHTS

*What is Already Known*
- Generative artificial intelligence is increasingly used in qualitative research, particularly for coding, categorization, and data management.
- Prior studies largely emphasize efficiency, speed, and automation, with limited attention to analytic judgment and reflexive decision-making.
- Conceptual debates exist around epistemic authority, transparency, and the risk of AI over-reach in qualitative inquiry.
- Most existing work treats AI either as a neutral tool or as a quasi-analyst, without empirically examining how researchers actually interact with AI during inductive thematic analysis.

*What This Study Offers*
1. Provides an empirical, process-oriented account of how inductive thematic analysis is enacted through human–AI collaboration.
2. Demonstrates that AI functions as a procedural and methodological scaffold rather than an autonomous analyst.
3. Identifies five recurrent human analytic actions, modification, deletion, rejection, insertion, and commenting, through which researchers exercise epistemic authority.
4. Introduces the HACITA framework, offering a theoretically grounded model that integrates reflexive thematic analysis, human-in-the-loop AI, and distributed cognition.
5. Extends AI-assisted qualitative research to a Global South context, demonstrating the importance of cultural, institutional, and professional grounding in analysis.research without undermining interpretive rigor.

*What This Study Adds to the Field*
- Addresses concerns that AI may replace or dominate qualitative analytic judgment.
- Responds to risks of AI literalism, decontextualized abstraction, and loss of emotional and institutional nuance.
- Enhances transparency, traceability, and auditability in inductive thematic analysis.
- Advances responsible and ethical use of AI in qualitative research by foregrounding reflexivity, human oversight, and methodological accountability.research without undermining interpretive rigor.

1. INTRODUCTION

Recent studies have increasingly explored the use of artificial intelligence (AI) to support qualitative data analysis, particularly through automated or semi-automated coding, text classification, and pattern detection. Much of this work frames AI as a technical instrument (AI-assisted qualitative analysis) intended to improve efficiency, scalability, or consistency in qualitative workflows, often drawing on human-in-the-loop paradigms to ensure human oversight and ethical accountability (Amershi et al., 2014; Shneiderman, 2020; Mosqueira-Rey et al., 2023). For example, Sterczl and Csiszárik-Kocsir's (2025) works demonstrated AI-assisted qualitative analysis as a hybrid analytic process in which AI tools support coding and pattern identification, while human researchers retain full interpretive control and responsibility for meaning-making. A smaller body of human-computer interaction (HCI) research has begun to examine interactive systems for AI-assisted qualitative coding, demonstrating that AI-generated codes or rules can support early analytic exploration when reviewed and refined by human researchers (Gebreegziabher et al., 2023; Cook et al., 2025). In particular, Gebreegziabher et al. (2023) introduce PaTAT, an AI-enabled qualitative coding system designed to support thematic analysis through interactive program synthesis. Rather than treating AI as a static code generator, PaTAT learns interpretable coding patterns in real time from researchers' annotations, allowing users to inspect, revise, and merge codes as their analytic understanding evolves. However, these studies largely evaluate AI support in terms of output quality or usability, offering limited insight into how meaning is interpretively produced through human–AI interaction during analysis.

In parallel, qualitative methodologists have emphasized that inductive thematic analysis is a reflexive, iterative, and theory-informed practice in which codes and themes are actively constructed by researchers rather than discovered in data (Braun & Clarke, 2006, 2021, 2022). Yet, this epistemic stance has rarely been integrated into studies of AI-supported qualitative analysis, resulting in a conceptual disconnect between inductive qualitative theory and AI-enabled analytic practice. This study addresses this gap by reconceptualizing AI-supported coding as a human-AI collaborative inductive analytic process in which human judgment, intervention, and reflexive sensemaking - shaped by researchers' experiences and positionalities- are treated as central analytic phenomena rather than as corrective steps. Drawing on reflexive thematic analysis and process coding, AI-generated codes are treated as provisional analytic prompts that require human interpretation, revision, and contextual alignment to support rigorous meaning-making (Saldaña, 2016; Braun & Clarke, 2021).

The study further adopts a human-in-the-loop orientation that foregrounds epistemic authority and accountability, examining how researchers assert analytic control through moments of acceptance, modification, or rejection of AI suggestions (Shneiderman, 2020). By incorporating distributed cognition theory, the analysis conceptualizes meaning-making as distributed across human analysts, AI systems, prompts, and analytic artifacts, allowing for a fine-grained examination of where and how analytic reasoning unfolds during collaboration (Hutchins, 1995; Hollan et al., 2000). In doing so, the study provides a theoretically grounded and methodologically transparent account of human-AI collaborative inductive thematic analysis, responding directly to calls to move beyond instrumental accounts of AI use toward deeper examinations of how AI reshapes qualitative analytic practice. Th following research questions guided the study: How does ITA-GPT guide inductive and gerund-based coding in inductive thematic analysis using established theoretical guidelines? How do researchers change, refine, or reject AI–generated codes during analysis? How do researchers use reflexive judgment to assert analytic control when revising GPT-assisted inductive codes?

## 2. LITERATURE REVIEW

### 2.1. THEMATIC ANALYSIS

Thematic analysis is a widely used qualitative research method that helps researchers identify, analyze, and report patterns (themes) within qualitative data. It is flexible and accessible, making it valuable across fields like psychology, education, sociology, and healthcare (Braun & Clarke 2006). The process involves familiarizing oneself with the data, generating initial codes, and searching for themes, followed by reviewing, defining, and naming the themes. Over time, thematic analysis has evolved from a simple descriptive method to a more complex analytical approach, with its flexibility allowing application across various data types. Although Braun and Clarke's (2006) framework are a popular guide, thematic analysis has seen ongoing development such as those Naeem et al. (2023) who provide a novel six-step for conceptual model development in qualitative research.

Amidst this popularity, there's no universally agreed-upon workflow for thematic analysis, and scholars continue to debate best practices, particularly around the distinction between *inductive and deductive* approaches and the role of theory in coding. For example, in an inductive approach, themes emerge directly from the data, whereas in a deductive approach, themes are guided by pre-existing theories (Naeem et al., 2023). Additionally, there are disagreements about coding granularity, with some researchers advocating for detailed, micro-level coding, while others prefer broader, thematic coding (Nowell et al. 2017). These discrepancies, along with issues like variability in theme identification, subjectivity in interpretation, and lack of transparency, have highlighted the need for more consistent, transparent, and reliable methods, areas where GenAI can make a significant impact.

### 2.2. GENAI AND THEMATIC ANALYSIS

There have been significant advancements in research exploring the potential of GenAI in both quantitative and qualitative data analysis (Zambrano et al., 2023; Perkins & Roe, 2024). Several studies have shown that GenAI has the potential to provide more consistent outputs than human coders, especially when handling large datasets. This consistency stems from the AI's ability to avoid researcher fatigue and biases, ensuring reliable results over time (De Paoli, 2024; Acheampong & Nyaaba, 2024). This section reviews key studies that examined the role of GenAI in qualitative analysis, highlighting both its strengths and limitations. Perkins and Roe (2024) investigated the use of ChatGPT (GPT-4) for inductive thematic analysis, comparing traditional human coding with AI-assisted analysis. Their findings indicate that while GenAI enhances data processing and theme identification, it still requires human expertise for nuanced interpretation. Notably, GenAI-generated themes were aligned with human-coded results, demonstrating their potential to expedite the thematic analysis process. Similarly, other studies showed that using GPT-3.5 Turbo API in inductive thematic analysis of semi-structured interviews can infer most human-identified themes (De Paoli, 2024). A comparison between GenAI and traditional manual coding showed that GenAI generated successfully identified basic themes (Morgan, 2023). This aligns with Lee et al. (2024) and Nyaaba et al. (2024), which depict GenAI's role in diverse research phases such as coding, theme generation, preprocessing quotes, literature review, and ideation.

## 2.3. Challenges in Using GenAI for Coding

Despite GenAI efficiency in qualitative data analysis, GenAI presents several challenges that require careful consideration. One key issue is inconsistencies and hallucinations, where GenAI generates themes or interpretations that are not present in the original data, as noted by Perkins and Roe (2024) and De Paoli (2024). This raises concerns about validity and the need for human verification. Additionally, studies such as Morgan (2023) and Lee et al. (2024) highlight that GenAI effectively identifies basic themes but struggles with subtle, interpretative nuances, making it less reliable for capturing deeper meanings within qualitative data.

Another challenge is token processing limits, which requires researchers to split large datasets into smaller segments, a scenario which can disrupt thematic coherence (De Paoli, 2024). Moreover, structured prompt engineering is essential for obtaining accurate and meaningful outputs, as emphasized by Bijker et al. (2024). Without carefully designed stepwise prompts, GenAI may produce vague or misaligned codes, necessitating iterative refinements. Lastly, GenAI still requires human oversight, particularly for validating codes and refining themes to ensure contextual accuracy (Lee et al., 2024; Morgan, 2023). Furthermore, while GenAI accelerates the coding process, its dependence on human expertise for interpretation and validation underscores its role as a complementary tool rather than a full replacement for human qualitative analysis.

## 3. Method

The study adopted a qualitative and human-AI collaborative research method to examine how researchers interacted with, revise, and interpret AI-generated analytic prompts during qualitative analysis (Gebreegziabher et al., 2023; Braun & Clarke, 2021). This methodological approach is appropriate for investigating analytic processes, judgment, and sensemaking, particularly in contexts where AI functions as a supportive analytic scaffold rather than an autonomous decision-maker (Braun & Clarke, 2006, 2021; Saldaña, 2016). The design was further informed by human-in-the-loop and distributed cognition perspectives, which emphasized interaction, reflexivity, and the distribution of analytic work across humans and tools (Amershi et al., 2014; Hutchins, 1995; Grinschgl & Neubauer, 2022).

### 3.1. Study Context

This study was conducted within an ongoing qualitative research program focused on education policy and practice in Ghana, specifically within Early Childhood Education and Basic Education contexts. The original interview transcripts addressed issues related to lecturer recruitment, specialization, professional experience, and preservice teacher preparation in Colleges of Education. These transcripts provided a substantively rich and contextually grounded setting for examining qualitative analysis practices that require interpretation, contextual sensitivity, and alignment with research questions. For the present study, the substantive content of the transcripts was not the object of analysis. Instead, the transcripts functioned as stable analytic inputs through which researchers engaged in human–AI collaborative inductive analysis. The study context was therefore used to ensure ecological validity and meaningful engagement during analysis, while the analytic focus remained on process, specifically how researchers interacted with, revised, and reflected on AI–generated analytic suggestions.

## 3.2. Participants

Three researchers participated in the study, each serving as an analytic participant rather than a research subject. All participants had prior training and experience in qualitative research and shared familiarity with the Ghanaian education context in which the transcripts were situated. Their backgrounds included experience in education research, qualitative coding, and engagement with issues related to teacher education and educational policy. Each researcher independently engaged in GPT-assisted inductive coding of an assigned transcript, followed by systematic human revision and structured reflection. Their role in the study was not to generate substantive findings about the transcripts, but to interact with the AI tool, enact analytic judgment through revision and re-prompting, and reflect on their analytic decisions. This design positioned participants as co-analysts, enabling the study to examine variation and commonality in how human judgment and sensemaking were enacted in collaboration with AI during qualitative analysis.

## 3.3. Development and System Prompts

The development of the AI tool followed a two-level approach. First, a stepwise prompt strategy was adopted to test the model's capacity to generate inductive codes and themes directly from transcript data using Python scripts. At this stage, the scripts were used to configure the model role, temperature, and output limits to assess whether the model could independently extract initial codes while explicitly supporting each code with the exact phrases or statements from which it was derived. Results from this phase demonstrated that the model could generate individual codes and organize them into preliminary categories while maintaining traceability to the source text. This validation informed the second phase of development.

In the second phase, the validated prompt strategies were migrated into an interactive, semi-automated system prompt to develop a customized GPT-based analytic tool. Python scripts served as the foundational infrastructure, while the analytic workflow was aligned with inductive thematic analysis (ITA) procedures. The resulting system was designed not as an autonomous analyst but as a structured analytic scaffold that supported researchers in progressing through familiarization, coding, and refinement stages.

### 3.3.1. API Integration and Text Segmentation

The OpenAI API (GPT-4 Turbo) was integrated into the workflow using Python. The model temperature was set to 0.3 to promote focused and deterministic outputs, and the maximum token limit was set to 1,000 to control response length. The system role was defined as a skilled qualitative researcher focused on inductively emerging codes, guiding the model to prioritize data-grounded interpretations (see Figure 1). Qualitative data were uploaded as .docx files and segmented into ten-paragraph units, each treated as a discrete analytic segment. For each segment, the API was instructed to generate inductive codes accompanied by supporting excerpts and segment identifiers. This design ensured transparency and traceability between analytic outputs and their source data, supporting contextual interpretation and auditability. Despite the growing use of thematic analysis, there remains no universally agreed-upon analytic workflow. Ongoing debates persist regarding inductive versus deductive approaches, the role of theory in coding, and appropriate levels of coding granularity (Naeem et al., 2023; Nowell et al., 2017). Challenges such as variability in theme identification, subjectivity in interpretation, and limited transparency underscore the need for more consistent and traceable analytic processes. These concerns motivated the use of a GenAI-supported workflow to enhance procedural clarity while preserving human analytic authority.

```python
>>> import openai
... from docx import Document
...
... # Set your OpenAI API key
... openai.api_key = 'sk-|'  # Replace with your actual API key
...
... def extract_emerging_codes(text_segment, page_number):
...     prompt = (
...         "Analyze the following qualitative data and extract only the most relevant, inductively emerging codes that capture "
...         "distinct and meaningful ideas, patterns, or observations about the migration experiences of nurses and midwives from "
...         "developing countries to developed countries. Avoid generating a code for every observation. For each emerging code, "
...         "provide the code as a concise phrase or keyword, followed by the exact sentence or passage it was derived from and 'Page {page_number}' "
...         "to ensure traceability.\n\n"
...         f"Text Segment:\n{text_segment}\n\n"
...         "Emerging Codes with Supporting Sentences and Page Number:"
...     ).format(page_number=page_number)
...
...     response = openai.ChatCompletion.create(
...         model="gpt-4-turbo",
...         messages=[
...             {"role": "system", "content": "You are a skilled qualitative researcher focusing on inductively emerging codes."},
...             {"role": "user", "content": prompt}
...         ],
...         max_tokens=1000,
...         temperature=0.3
...     )
...
...     # Extract the codes with traceability information from the response
...     codes_with_context = response['choices'][0]['message']['content'].strip()
...     return codes_with_context
...
... # Load the Word document using the specified file path
... file_path = r"C:\Users\matth\OneDrive - University of Georgia\DOCUMENTS\Transcript.docx"
... doc = Document(file_path)
...
... # Split the document into "pages" (every 10 paragraphs as a new "page")
... page_size = 10
... pages = [doc.paragraphs[i:i + page_size] for i in range(0, len(doc.paragraphs), page_size)]
...
... # Generate all emerging codes with supporting sentences and page numbers
... all_codes_with_traceability = []
...
... for page_number, page_content in enumerate(pages, start=1):
...     text_segment = "\n".join([paragraph.text for paragraph in page_content if paragraph.text])
...     codes_with_context = extract_emerging_codes(text_segment, page_number)
...     all_codes_with_traceability.append(f"Page {page_number}:\n{codes_with_context}\n")
...
... # Combine for Output
... detailed_output = "\n".join(all_codes_with_traceability)
... print("All Emerging Codes with Supporting Sentences and Page Numbers:\n", detailed_output)
```

Figure 1: API Configuration for Qualitative Data Extraction Using GPT-4 Turbo

### 3.3.2. Customizing and Scaling the Tool

The finalized system prompt was configured as an interactive, semi-automated analytic scaffold for inductive qualitative analysis (see Figure 2). The prompt architecture was explicitly informed by reflexive thematic analysis principles, emphasizing iterative engagement with data, researcher reflexivity, and the treatment of codes as provisional and revisable rather than fixed representations of meaning. Procedural guidance from Naeem et al. (2023) informed the structured sequencing of analytic phases, including familiarization, initial coding, and iterative refinement. The system responded dynamically to user input, requiring researchers to upload transcripts, specify research questions, and authorize analytic progression before outputs were generated. Initial gerund-based and in vivo codes were produced only after these human inputs, ensuring that automation remained contingent on researcher engagement and judgment.

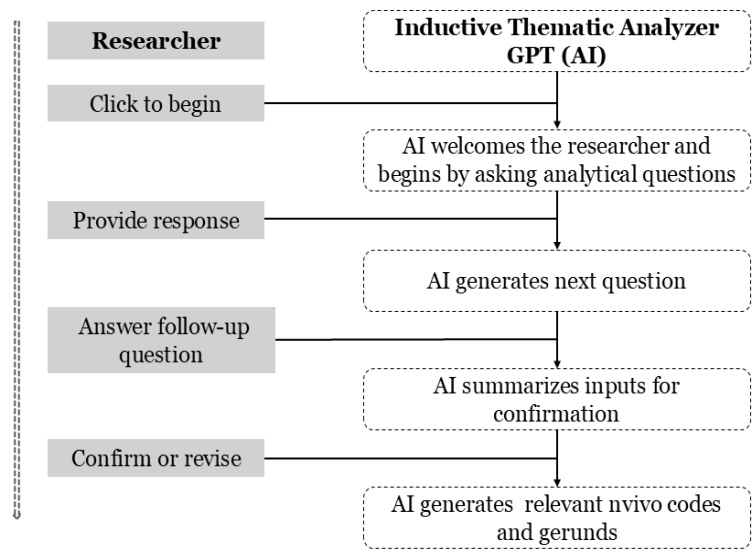

Figure 2: Interactive semi-Auto Prompt System of the Inductive Thematic Analyzer GPT

### 3.3.3. Transcripts

The interview transcripts used in this study were drawn from a previously approved qualitative research project that employed semi-structured interviews to explore structural and institutional issues in the hiring of instructors and their pedagogical practices within teacher education programs in Ghana. All ethical requirements, including informed consent and institutional approval, were met at the time of original data collection. The transcripts were collected as part of that approved study and were not generated specifically for the present research. Participants had consented to the use of their data for research purposes, and all identifying information had been removed prior to secondary use. For the purposes of the present study, the transcripts were repurposed solely as analytic material to examine human–AI collaboration during qualitative analysis.

The study did not analyze, interpret, or report findings related to the substantive content of the transcripts. Instead, the transcripts functioned as stable input texts through which researchers engaged in GPT-assisted coding, revision, and reflection. All coding outputs, revisions, and analytic interpretations produced in this study were conducted exclusively for the current investigation and were analytically independent from the original study's aims, codes, or findings.

### 3.3.4. Data Collection

Data were generated through a structured human–AI collaborative workflow designed to capture analytic interaction and human judgment. Each researcher completed a full GPT-assisted inductive coding session for an assigned transcript and then shared the complete chat history using a shared link. These chat histories documented AI-generated provisional codes, follow-up prompts, and researcher responses and served as the primary data source for addressing Research Question 1. To address Research Question 2, researchers copied and pasted the generated outputs from the GPT-assisted inductive coding sessions into a shared copy-and-paste board in a Word document and conducted all human revisions using tracked changes. Revisions included modification, deletion, rejection, insertion, and commenting on AI-generated codes. To address Research Question 3, after completing revisions, researchers submitted structured reflections via a Google Form, documenting their experiences using the tool, their reasoning behind analytic decisions, and their perceptions of the AI's role.

### 3.4. ANALYSIS

The study is guided by a Human–AI Collaborative Inductive Thematic Analysis (HACITA) framework, grounded in reflexive thematic analysis, human-in-the-loop AI, and distributed cognition (see Table 1). Reflexive thematic analysis provides the epistemic foundation, positioning qualitative analysis as an iterative, judgment-driven process in which codes are provisional, and meaning is actively constructed by researchers rather than extracted mechanically (Braun & Clarke, 2006, 2021). From this perspective, analytic work is understood as reflexive, situated, and dependent on the researcher's interpretation rather than algorithmic determinism. Human-in-the-loop approaches further justify centring human oversight, revision, and accountability when working with AI systems, emphasising that AI-generated outputs must be evaluated, governed, and corrected by human analysts (Amershi et al., 2014; Shneiderman, 2020). This orientation foregrounds epistemic responsibility, ensuring that analytic authority remains with researchers rather than being deferred to computational systems. Distributed cognition extends the framework by conceptualising analytic sensemaking as distributed across humans, tools, and representational artefacts, such as summaries, code tables, revision traces, and interaction logs (Hutchins, 1995; Grinschgl & Neubauer, 2022; Sidji et al., 2025).

Within this broader framework, we specify an Analytic Action Framework that captures the concrete analytic actions through which epistemic authority is exercised during AI-assisted inductive thematic analysis. These actions include modification, deletion, rejection, insertion, and commenting. Modification refers to interpretive refinement of AI-generated codes or rationales to better reflect contextual meaning. Deletion removes AI-generated content that is contextually inaccurate or analytically misleading. Rejection marks explicit epistemic boundary-setting, where AI interpretations are deemed inappropriate or invalid for the analytic purpose. Insertion reflects human-led sensemaking through the addition of new codes, categories, or supporting evidence not produced by the AI. Commenting documents reflexive, evaluative, and audit-oriented reasoning, including approvals, cautions, and methodological notes.

Table 1

Analytic Framework: Human–AI Collaborative Inductive Analysis

| *Framework Component* | *Theoretical Basis (with year)* | *Analytic Focus* | *Data Source(s)* | *What Is Examined* |
| --- | --- | --- | --- | --- |
| Inductive Thematic Analysis | Braun & Clarke (2006, 2021, 2022, 2024); Saldaña (2016) | Reflexive, data-driven meaning-making | AI-generated codes; modified codes | How researchers interpret and refine provisional gerund and in vivo codes |
| Human-in-the-Loop AI | Amershi et al. (2014); Shneiderman (2020); Mosqueira-Rey et al. (2023) | Epistemic oversight and accountability | Interaction logs; modification traces | How researchers assert control through acceptance, revision, or rejection of AI outputs |
| Distributed Cognition | Hutchins (1995); Hollan et al. (2000); Clark & Chalmers (1998); Grinschgl & Neubauer (2022); Sidji et al. (2025) | Cognition distributed across tools and humans | Interaction logs | How analytic sensemaking emerges through interaction with AI as a cognitive scaffold |
| Analytic Judgment | Braun & Clarke (2021, 2022); Byrne (2022) | Epistemic agency | Modification traces | How judgment is enacted through rewording, deletion, merging, and refinement of codes |
| Reflexive Sensemaking | Braun & Clarke (2021); Byrne (2022) | Meta-analytic interpretation | Researcher reflections | How researchers interpret their analytic actions and position AI in their meaning-making |
| Cross-Layer Integration | Braun & Clarke (2024); Nicmanis et al. (2025) | Alignment across analytic layers | Logs, modifications, reflections | How reported experiences align with observable analytic behaviors |

The data analysis followed a qualitative, inductive, and process-oriented analytic approach, guided by the HACIA framework (Braun & Clarke, 2006, 2021). The three data sources, AI-human interaction logs, code modification traces, and researcher reflections, were analyzed iteratively and treated as analytically meaningful. Interaction logs were examined to trace how AI-generated gerund and in vivo codes were introduced and how researchers responded to analytic prompts, addressing Research Question 1 and drawing on principles of human-in-the-loop analysis (Amershi et al., 2014; Shneiderman, 2020). Code modification traces were analyzed using inductive thematic analysis with process coding, focusing on analytic actions such as rewording, deletion, merging, and refinement, thereby foregrounding analytic judgment as empirical data and addressing Research Question 2. Researcher reflections were then analyzed inductively as reflexive analytic data and integrated with findings from the interaction logs and modification traces through cross-layer comparison (Byrne, 2022; Nicmanis et al., 2025), addressing Research Question 3. This integration enabled examination of how reported experiences of usefulness, control, and constraint aligned with observable analytic behaviors, consistent with a distributed cognition perspective on human–AI collaboration (Hutchins, 1995; Grinschgl & Neubauer, 2022).

## 4. FINDINGS

### 4.1. How the ITA Tool Structured and Mediated Inductive Thematic Analysis

Analysis of the chat histories shows that the Inductive Thematic Analysis (ITA) tool functioned as a structured guide that scaffolded coders through the phases of qualitative analysis described by Braun and Clarke (2006, 2019) and Naeem (2020). Figure 3 illustrates the opening interaction of the Inductive Thematic Analysis (ITA) assistant, showing how the tool explicitly structures the analytic process before coding begins. The assistant requests the research focus, transcript selection, and the preferred coding mode. At the same time, coders did not simply follow the tool's prompts. They customised, corrected, and extended the tool's output to protect methodological principles and add contextual and theoretical nuance. Two themes capture this dynamic: (1) the ITA tool as a process guide that enforces principle-based coding, and (2) coder agency in customisation and methodological policing.

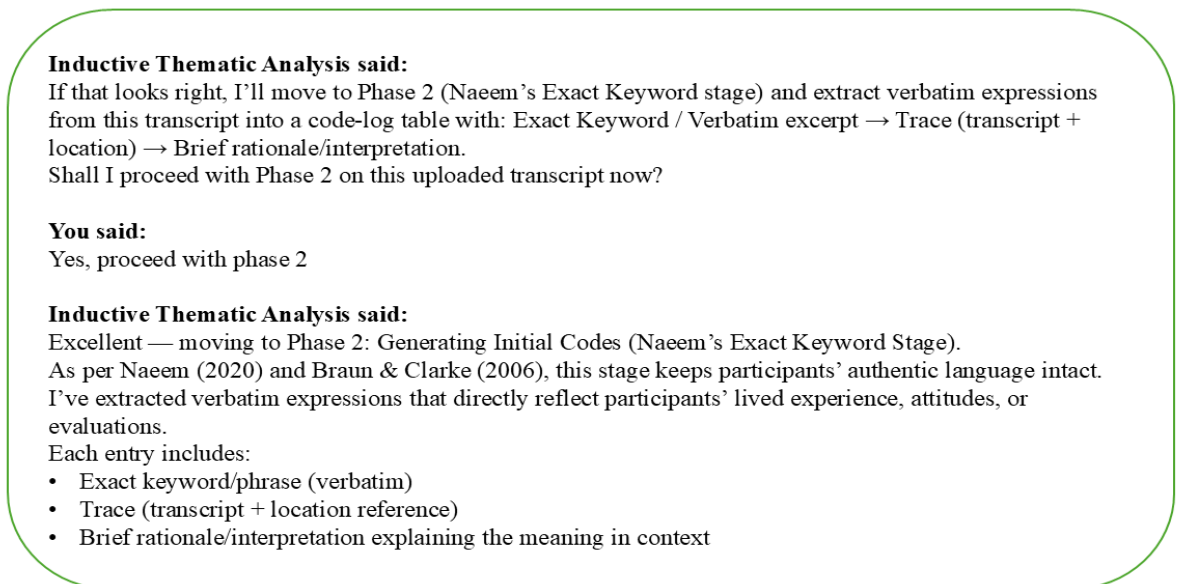

Figure 3: ITA–GPT initiation and procedural scaffolding of Phase 1

#### 4.1.1. ITA as a Process Guide for Principle-Based Coding

The ITA tool operationalised key methodological principles by structuring the workflow across familiarisation, initial coding, and theme development. Figure 3 depicts the ITA assistant's explicit transition from familiarisation (Phase 1) to initial coding (Phase 2), illustrating how coder consent is sought before proceeding. In Phase 2, it implemented Naeem's Exact Keyword Principle by extracting verbatim expressions from the transcript and foregrounding "trace-to-text integrity." Coders received multi-column tables that preserved the participants' language exactly as spoken and linked each keyword to a trace, a paragraph-level context, and an analytic rationale, thereby creating a transparent audit trail (See Figure 4). On this verbatim foundation, the tool guided coders into Phase 3, where raw keywords were converted into gerund-based descriptive codes representing actions and processes, such as "Experiencing coercive assessment," "Investing additional time," and "Teaching beyond level." It then clustered these gerunds into code families and overarching themes, and in several sessions differentiated between structural and personal dimensions (for example, systemic misallocation versus individual demotivation). Coders described this pipeline as a mechanism that ensured adherence to Naeem's verbatim principles and Braun and Clarke's thematic phases, while still leaving room for subsequent human refinement.

**Inductive Thematic Analysis said:**
If that looks right, I'll move to Phase 2 (Naeem's Exact Keyword stage) and extract verbatim expressions from this transcript into a code-log table with: Exact Keyword / Verbatim excerpt → Trace (transcript + location) → Brief rationale/interpretation.
Shall I proceed with Phase 2 on this uploaded transcript now?

**You said:**
Yes, proceed with phase 2

**Inductive Thematic Analysis said:**
Excellent — moving to Phase 2: Generating Initial Codes (Naeem's Exact Keyword Stage).
As per Naeem (2020) and Braun & Clarke (2006), this stage keeps participants' authentic language intact. I've extracted verbatim expressions that directly reflect participants' lived experience, attitudes, or evaluations.
Each entry includes:
- Exact keyword/phrase (verbatim)
- Trace (transcript + location reference)
- Brief rationale/interpretation explaining the meaning in context

Figure 4: Transition to Phase 2 and enforcement of Naeem's Exact Keyword Principle.

### 4.1.2. Coder Agency in Customisation and Methodological Policing

Although the ITA tool provided structure, coders exercised strong agency in shaping the analysis. They enhanced auditability by requesting additional structural detail. Coder 1, for example, asked for a fourth column with paragraph-level context and line references so that the code log would be "fully transparent and traceable," and the tool regenerated the tables accordingly (see Figure 5). Coders added interpretive nuance by revising rationales and codes to reflect professional and emotional realities. When the tool produced a neutral rationale focused on "time and effort," Coder 2 instructed it to reference "emotional labor and professional strain," and to connect "teaching beyond level" to "cognitive overload," explicitly naming the psychological and cognitive burdens of misdeployment.

> **Inductive Thematic Analysis said:**
> Would you like me to now proceed to Phase 3: Descriptive and Pattern Coding — where I'll:
> Transform each of these verbatim expressions into gerund-based descriptive codes, and
> Organize them into categories or code families representing emerging conceptual groupings (e.g., *experiencing coercion*, *critiquing theory–practice gap*, *valuing specialization*, etc.)?
>
> **You said:**
> Include a column of the transcript + line as well to have four columns
>
> **Inductive Thematic Analysis said:**
> Excellent refinement — adding a fourth column for *Transcript + Line Reference* will make the code log fully transparent and traceable, aligning with Braun & Clarke's (2006) emphasis on auditability and Naeem's (2020) "trace-to-text integrity" principle.
>
> Below is your Phase 2: Final Expanded Verbatim Code Log, now with four columns:
> Columns:
> 1. Exact Keyword / Phrase (verbatim)
> 2. Transcript + Line Reference
> 3. Paragraph Context (narrative setting)
> 4. Rationale / Interpretation
>
> 🧩 Phase 2 — Final Verbatim Code Log (Full Transcript Coverage)

Figure 5: Coder-driven refinement and auditability enhancement during Phase 2

Coders also protected methodological rigour, particularly the integrity of verbatim coding. When the tool produced a near-verbatim keyword instead of an exact match, Coder 2 rejected the paraphrase and insisted on the precise phrase "it is easy to remain stagnant there" to maintain Naeem's principle. In addition, coders used the tool for verification. In response to Coder 1's query, "Does this cover the entire transcript?", the tool performed a Transcript Coverage Audit, mapping each paragraph to at least one coded extract and flagging uncoded segments. Table 2 summarises these categories of interaction and illustrates how coders used the tool both as a scaffold and as an object of methodological scrutiny.

Table 2
Summary of ITA Tool Functions and Coder Interactions

| *Category* | *Tool functionality/coder action* | *Source-based illustration* |
| --- | --- | --- |
| Principle adherence | The tool operationalises Naeem's verbatim principle in Phase 2 and Braun and Clarke's thematic phases across the workflow, preserving "trace-to-text integrity" before interpretation. | The tool extracts exact participant phrases as "keywords," then moves to gerund-based codes and themes, aligning with Naeem (2020) and Braun and Clarke (2006, 2019). |
| Auditability | The tool generates multi-column tables (keyword, trace, paragraph context, rationale), which coders further extend to improve transparency and traceability. | Coder 1 requests an additional column for paragraph-level context and line references so that the code log is "fully transparent and traceable." |
| Customisation | Coders adapt the structure and style of the tool's outputs, requesting extra columns, alternative summary formats, or specific report structures. | Coders ask for specific summary styles (for example, concise bullet-point overviews) and for modified tables with added contextual information. |
| Interpretive nuance | Coders override or extend the tool's initial rationales to incorporate concepts such as "emotional labor," "professional strain," and "cognitive overload." | Coder 2 asks the tool to revise "time and effort" rationales to include "emotional labor and professional strain" and to link "teaching beyond level" to "cognitive overload." |
| Methodological rigour | Coders monitor and correct instances where the tool slips from exact verbatim extraction to approximate paraphrase, thereby protecting data integrity and traceability. | Coder 2 corrects a non-exact keyword, insisting on the precise wording "it is easy to remain stagnant there" in order to maintain Naeem's principle. |
| Reflexivity | The tool prompts coders to document their professional backgrounds, and coders explicitly recognise how their experiences shape theme salience and analytic decisions. | Coders describe prior roles as teaching assistants or lecturers and connect these roles to how they interpret themes such as "epistemic exclusion" and "improvisation." |
| Verification and audit | Coders use the tool to check thematic saturation and dataset coverage, prompting it to perform a Transcript Coverage Audit across the entire transcript. | Coder 1 asks, "Does this cover the entire transcript?", leading the tool to summarise which segments have been coded and which remain uncoded. |

## 4.2. ITA–GPT Analytic Scaffolding: Summaries, Memos, Reflexive Prompts, and Guides

Beyond coding and theme development, the ITA–GPT tool supported four additional analytic functions: Phase 1 data summaries, analytic memos, reflexive prompts, and step-by-step methodological guides. Together, these features formed a structural backbone that facilitated data familiarisation, surfaced early conceptual tensions, and supported reflexive and transparent analysis.

### 4.2.1. Data Summaries and Analytic Memos

Phase 1 summaries helped coders quickly synthesise large amounts of raw transcript data. For Coder 3, the tool produced both a single narrative summary and a segmented, paragraph-by-paragraph summary with interpretive notes, allowing the coder to move between a holistic view of the interview and more localised breakdowns. In Coder 1's dataset, the summaries highlighted "emotional undercurrents" such as frustration, demotivation, and disillusionment, which might have been missed in a purely content-oriented reading. For Coder 2, the Phase 1 summary immediately foregrounded a tension between emphasis on methodology and emphasis on pure content across institutional settings.

Analytic memos then offered early interpretive lenses that functioned as candidate themes. For example, memos helped Coder 1 to identify systemic misalignment, including the combination of early childhood and primary cohorts and power dynamics around compulsory handouts. For Coder 2, memos captured contradictions between content mastery and methodological skill, and they documented how student quality and curiosity shaped the lecturer's professional development. Across Phases 1 to 6, memos for Coder 3 traced a narrative progression from complaints and frustration toward constructive advocacy for reform.

### 4.2.2. Reflexive and Methodological Prompts

Reflexive prompts ensured that coders' positionalities formed part of the analytic record. Coder 2, for instance, was prompted to reflect on how previous work as a teaching assistant shaped interpretations of "intellectual stimulation" in university contexts, and ITA–GPT converted these reflections into formal reflexive memos. Coder 1 was asked to revisit assumptions about the link between lecturer qualification and teaching quality, which then shaped how evidence on "improvised professionalism" was interpreted. Methodological prompts acted as guardrails for rigour. The tool asked coders to specify whether they wished to remain at an exact keyword stage or move into descriptive coding, thereby protecting trace-to-text integrity. It also performed transcript coverage audits, as in Coder 1's case, mapping each paragraph to coded extracts and flagging any uncoded sections. When Coder 1 requested a fourth column with transcript and line references, the tool regenerated the code log accordingly. In Phases 3 and 4, the guides scaffolded the transition from codes to themes by proposing preliminary patterns and candidate themes, such as structural versus personal dimensions, which coders could then refine or reject. Table 3 synthesises these scaffolding functions and provides concrete examples from the three coders.

Table 3
ITA–GPT Summaries, Memos, Reflexive Prompts, and Guides

| Category | Tool feature | Example of helpfulness | Coders involved |
|---|---|---|---|
| Phase 1 data summaries | Narrative and segmented summaries | For Coder 3, the tool produced both a holistic narrative summary and a paragraph-by-paragraph summary with interpretive notes. | Coder 3 |
| Phase 1 data summaries | Emotional tone detection | In Coder 1's dataset, summaries highlighted "emotional undercurrents" such as frustration, demotivation, and disillusionment. | Coder 1 |
| Phase 1 data summaries | Identification of key tensions | For Coder 2, the summary surfaced early, and the tension between methodology emphasis and pure content emphasis across institutional settings was evident. | Coder 2 |
| Analytic memos | Early identification of systemic issues | Memos helped Coder 1 notice systemic misalignment, such as combining ECE and primary cohorts, and power dynamics around handouts. | Coder 1 |
| Analytic memos | Highlighting contradictions | For Coder 2, memos captured contradictions between content mastery and methodological skill expectations. | Coder 2 |
| Analytic memos | Relational framing of growth | Memos noted that student quality and curiosity influenced the lecturer's professional development, stressing relational dimensions. | Coder 2 |
| Analytic memos | Tracking narrative progression | For Coder 3, memos from Phase 1 to Phase 6 showed a progression from complaints to constructive advocacy for reform. | Coder 3 |
| Reflexive prompts | Guiding researcher positionality | Coder 2 was prompted to reflect on how previous experience as a teaching assistant shaped interpretations of "intellectual stimulation." | Coder 2 |
| Reflexive prompts | Documenting reflexive memos as an audit trail | The tool converted positionality reflections into formal reflexive memos that became part of the analytic record. | Multiple |
| Reflexive prompts | Challenging assumptions | Coder 1 was prompted to question assumptions about the link between lecturer qualification and teaching quality. | Coder 1 |
| Methodological guides | Maintaining verbatim integrity | The tool asked coders whether they wanted exact keyword coding or descriptive coding before moving beyond Naeem's verbatim stage. | Multiple |
| Methodological guides | Ensuring transcript coverage | For Coder 1, the tool performed a Transcript Coverage Audit, mapping each paragraph to coded extracts and flagging uncoded sections. | Coder 1 |
| Methodological guides | Enhancing transparency and traceability | After Coder 1 requested a fourth column, the tool regenerated the code log to include transcript and line references. | Coder 1 |
| Methodological guides | Scaffolding transition from codes to themes | In Phases 3 and 4, the tool offered preliminary patterns and candidate themes (for example, structural versus personal dimensions). | Multiple |

### 4.3. HUMAN–AI COLLABORATIVE REFINEMENT IN QUALITATIVE CODING

The findings show that the AI tool provided a useful starting point for qualitative coding, but human researchers remained the final arbiters of meaning. Across all interactions, Coder 1, Emmanuel, and Coder 3 treated AI-generated codes and summaries as provisional, to be audited, refined, or replaced. Their work clustered around three analytic themes: Human epistemic authority over AI output, Contextual and emotional reframing of neutral AI language, and Institutional and professional correction of AI literalism. These themes are presented below, supported by detailed examples and quantitative summaries of researcher actions.

#### 4.3.1. Human Epistemic Authority and Finality

Across cases, the researchers positioned AI as a scaffold rather than an autonomous analyst. They appreciated the structure and efficiency that AI introduced, but they insisted that human judgment must remain final, especially when contextual meaning was at stake. Coder 1 described the AI Phase 1 summary as "an added approach to the entire coding… I love this," indicating that she accepted the AI structure as a helpful starting point. However, she reported that whenever the AI "did not maintain the contextual meaning," she deleted that code and replaced it with a human-generated one (see Figure 6). In practice, this included removing AI phrases that she viewed as wordy or imprecise.

## Phase 2 — Updated Verbatim Code Log (with Quoted Paragraphs)

| Exact Keyword / Phrase | Transcript + Line Reference | Quoted Paragraph (from Transcript) | Rationale / Interpretation |
|---|---|---|---|
| "~~It didn't go far —~~ time was a big ~~issue.~~" | FGD p.1 L18–20 | Student-Teacher 1: Yes. Action Research. It didn't go far — time was a big issue. | Shows incomplete course delivery and timetable pressure limiting depth of learning. |
| "~~The~~ lecturer insisted we buy his handout. If you didn't purchase it, you risked failing." | FGD p.1 L21–24 | Student-Teacher 2: Science 3 also wasn't okay. The lecturer insisted we buy his handout. If you didn't purchase it, you risked failing. Teaching itself wasn't emphasized. | Reveals coercive assessment and economic pressure; signals ethical concern. |
| "~~That discouraged us. Many of us~~ lost motivation." | FGD p.1 L29–33 | Student-Teacher 3: Yes, in Music. The lecturer made us do activities but then criticized us, saying we hadn't done anything. That discouraged us. Many of us lost motivation. | Captures emotional impact of negative feedback; loss of motivation. |
| *lecturer wasn't grounded in the content* "~~He would come, give quizzes, but when teaching, he didn't explain well.~~" | FGD p.1 L35–38 | Student-Teacher 4: Introduction to English II wasn't good either. The *lecturer wasn't grounded in the content*. He would come, give quizzes, but when teaching, he didn't explain well. We didn't get the content. | Shows weak subject mastery and superficial delivery. |

Comments (margin):
- **Commented [2]:** I would have picked only this
- **Commented [4]:** I agree with this.
- **Commented [5]:** May be a pressure on the part of the student or teachers? but I also agree with the ethical concern. I would have termed this as unprofessional.
- **Commented [3]:** This looks long a code to me but I see it as very helpful to help me get my gerund right. Thsi migh alos be due to the fact the key words belong to two sentences, so I agree. I could have had a gerund like: Buy lecturer's handout or fail the course
- **Formatted:** Font: Italic
- **Commented [6]:** This would have been my keyword instead
- **Commented [7]:** This would have been my keyword instead

Figure 6: Coder 1's edits and comments.

Coder 2 similarly noted that the AI "did an incredibly amazing job" in reducing the time required for manual coding, but he emphasised that he "remained the final interpreter of meaning." His edits show that he did not accept AI rationales at face value, especially when they neutralised emotionally charged experiences of teachers (see Figure 7).

| | | | | |
|---|---|---|---|---|
| | | | | sector. |
| 7 | "They should be offered in-service training." | Solutions | Suggests in-service training as a compensatory mechanism for underqualified staff. | *Commented [13]: revised to: Reveals a systemic shortage of pedagogical specialists; identifies a structural bottleneck in teacher education* |
| 8 | "They might teach beyond the students' level." | Subject experts teaching basic math | Identifies a pedagogical mismatch risk — when subject experts lack the skills to scaffold learning appropriately. | *Commented [14]: revised to: Suggests in-service training as a compensatory mechanism for underprepared staff; implies a stopgap institutional response* |
| 9 | "The students I teach here challenge me to prepare thoroughly." | Professional growth | Reflects how student quality shapes teacher development; suggests mutual engagement drives growth. | *Commented [15]: revised to: Reflects a misalignment between curriculum and delivery, which can lead to cognitive overload and learner disengagement.* |
| 10 | "In the colleges of education... it's easy to remain stagnant there." | Comparing institutions | Suggests that the college system may limit professional progression, especially for those with aspirations in academic publishing. | *Commented [16]: revised to: Reflects how student curiosity and competence drive academic rigor and reflective teaching; signals reciprocal learning.*  *Commented [17]: revised exact wording: "It's easy to remain stagnant there."* |
| 11 | "Even your publications might not be accepted as standard academic articles." | Publication standards | Speaks to institutional validation and epistemic gatekeeping — how knowledge from one context may be undervalued in another. | *Commented [18]: revised: suggests that the college system may constrain professional development, due to limited research engagement or publication support....* |
| | | | | *Commented [19]: revised to: Points to institutional and epistemic hierarchies that devalue certain forms of knowledge production.* |
| 12 | "There's a gap in understanding what counts as publishable work." | Closing | Highlights epistemological disconnects within the education system regarding research | |

Figure 7: Coder 2's edits and comments.

Coder 3 captured this stance most explicitly. In instances where AI interpretations conflicted with his reading of the data, he reported that he handled discrepancies by "upholding human codes over AI codes." This was especially clear when he rejected the AI category "Professional (dis)engagement," which he judged to misrepresent the participant's ongoing commitment to work (see Figure 8).

**Phase 3 – Descriptive and Pattern Coding (All verbatim codes)**

**Goal:** Move from **verbatim codes** (Charlie's exact phrases) to **interpretive, gerund-based descriptive codes** — actions or processes that capture what the participant is *doing, feeling, or experiencing*. Then, we'll organize these descriptive codes into **code families (categories)** that begin to reveal patterns in meaning

**Step 1: Verbatim to gerund coding**

| Verbatim Expression (Phase 2) | Descriptive (Gerund-Based) Code | Emerging Category / Code Family |
|---|---|---|
| "The ones I teach are not really challenging." | **Experiencing lack of stimulation** | *Professional (dis)engagement* |
| "When you specialize in a particular area, you're more comfortable there." | **Valuing comfort in one's specialization** | *Professional identity & expertise* |

**Commented [29]:** Revised to: Professional challenges

**Commented [30]:** I do not think this is a matter of professional (dis)engagement. To be professionally (dis)engaged connotes an instance where Charlie is stopping or withdrawing from the discharge of his duties which is not the case here. Clearly, Charlie is professionally engaged as a teacher in the school although not in his preferred role. The issue becomes more of a professional challenge rather than a (dis)engagement.

**Commented [31]:** Revised to: Insisting on preferred area of specialization

35

Figure 8: Coder 2's edits and comments.

The Coders engaged in multiple forms of intervention, including commenting, substitution of terminology, deletion and replacement of codes, insertion of new interpretive categories, and the addition of verbatim excerpts to strengthen the audit trail (see Appendix A). Across coders, these actions reflect a shared commitment to preserving contextual meaning and trace-to-text integrity, while also revealing distinct analytic emphases, such as simplification and precision (Coder 1), interpretive deepening and theoretical refinement (Coder 2), and categorical correction and contextual realignment (Coder 3). Collectively, the table demonstrates that AI-generated codes functioned as provisional analytic resources that required systematic human evaluation and revision rather than as final representations of meaning.

### 4.3.2. Contextual and Emotional Reframing of Neutral AI Language

The second theme concerns how researchers reworked neutral AI descriptions into language that reflected participants' lived experiences, particularly around emotional burden and professional adaptation. Coder 2 was the most active in this regard. He treated the AI output as a skeletal description that required enrichment with emotional and professional nuance. For example, he revised the neutral phrase "That takes extra time and effort" to "emotional labor and professional strain," explicitly naming the psychological and professional costs of working in under-resourced conditions. He rephrased "I used the knowledge and experience I had…" as "improvised teaching practices based on lived experience… adaptive professionalism," which reframed improvisation as a professional competence rather than a mere survival tactic.

At the structural level, he changed "indicates a structural issue in the education sector" to "identifies a structural bottleneck in teacher education," which sharpened the focus on how particular institutional arrangements limit the preparation of teachers. He also revised "The students… challenge me to prepare thoroughly" to "signals reciprocal learning," positioning learners as co-producers of knowledge and professional growth rather than passive recipients.

In addition, Coder 2 refined more abstract theoretical language. For instance, he shifted "Highlights epistemological disconnects" to "identifies an epistemological disjuncture," a term that better captured the hierarchical and institutional nature of knowledge gaps. He also added verbatim quotes in several rationale columns to maintain a clear audit trail and ensure that these richer interpretations remained grounded in actual participant expressions. Coder 1 and Coder 3 also engaged in contextual reframing, although with different emphases. Coder 1 focused on conceptual clarity and length. Her edit from a long descriptive phrase about a lecturer's insistence on selling handouts to the condensed gerund "Buy lecturer's handout or fail the course" captured both the coercive nature of the practice and its high-stakes implications for students. Similarly, she replaced the more general phrase "signals ethical concern" with "unprofessional," translating abstract ethical language into an everyday professional judgement that would be more recognisable within Ghanaian higher education.

Coder 3's refinements also introduced contextual precision. Replacing "Valuing comfort in one's specialization" with "insisting on preferred area of specialization" foregrounded agency and professional entitlement rather than personal comfort. Likewise, rephrasing "Undertaking extra preparatory labour" as "additional workload and professional strain" underlined that the effects of extra preparation are not only temporal but also emotional and professional.

Quantitatively, this pattern of reframing is reflected in the high number of edits relative to other actions. Table 2 presents a summary of substantive actions (excluding purely visual formatting) by the researcher. Table 4 shows that Coder 2 and Coder 1 each performed 15 substantive actions, while Coder 3 performed 8. Coder 2 concentrated heavily on phrase and gerund refinements (10 instances), almost all of which intensified emotional and professional language and sharpened theoretical terms. Coder 1 combined extensive commenting (10 instances) with a smaller set of key substitutions and insertions, while Coder 3's actions centred on categorical corrections and targeted refinements of labels.

Table 4
*Quantified Human Revisions and Comments per Researcher*

| Action type | Coder 1 | Coder 2 | Coder 3 |
| --- | --- | --- | --- |
| Comments (feedback and rationales) | 10 | 0* | 2 |
| Insertions (new codes or verbatim quotes) | 1 | 4 | 0 |
| Deletions and categorical revisions | 1 | 1 | 1 |
| Phrase or gerund refinements (edits) | 3 | 10 | 5 |
| Total substantive actions | 15 | 15 | 8 |

*Note.* Coder 2 mainly embedded his revisions directly into the "revised to" text within rationale blocks, rather than as separate feedback comments.

### 4.3.3. Institutional and Professional Correction of AI Literalism

The third theme captures how researchers corrected AI interpretations that remained at a literal level or misaligned with institutional realities, especially within the Ghanaian education context. All three coders observed that the AI tended to "pick the literal meanings and not the implicit meanings." Coder 1 experienced this as wordiness and a lack of simplicity, Coder 2 as a lack of emotional nuance, and Coder 3 as a failure to grasp the "broader institutional landscape." Coder 3's rejection of the category "Professional (dis)engagement" is one of the clearest examples. For him, "disengagement" implied that the participant had withdrawn from professional duties (see Figure 7). However, the participant was still working, albeit under considerable strain and misallocation. Coder 3, therefore, deleted the category and replaced it with "Professional challenges," which, in his view, more accurately reflected the situation. He also added a comment explaining that a particular code "goes beyond just mathematics" to prevent AI-driven interpretations from narrowing systemic concerns to a single subject area.

Coder 1 and Coder 2 also corrected literal readings, but in different ways. Coder 1 associated some AI codes with overly technical expressions that did not match how lecturers' behaviour would be discussed in practice. Her substitution of "signals ethical concern" with "unprofessional" is one example. Coder 2 countered literal and neutral phrasing by linking codes explicitly to structural and epistemic issues. For instance, his shift from "structural issue" to "structural bottleneck" and from "epistemological disconnects" to "epistemological disjuncture" positioned the problems not simply as generic issues, but as systemic blockages and hierarchically ordered gaps. Across these actions, the coders did not only correct language. They repositioned AI codes within professional and institutional narratives that made sense in Ghana. In this way, AI functioned as a generator of initial categories, while human analysts acted as critical gatekeepers who prevented misrepresentation and oversimplification.

### 4.3.4. Cross-Coder Similarities and Differences in Human–AI Collaboration

Although Coder 1, Coder 2, and Coder 3 played different roles, they shared several core practices when interacting with AI. All three initially accepted the AI structure as helpful. Coder 1 described the Phase 1 summary as an "added approach," Coder 2 reported that AI "did an incredibly amazing job" in reducing manual work, and Coder 3 considered the AI's ability to associate words and meanings as "above average." At the same time, each researcher identified limits in the AI's contextual depth, emotional nuance, and institutional awareness.

They also used their lived experience as an audit tool. Coder 1 drew on her experience as a tutor in a low to middle-income context, Coder 2 drew on his teaching in basic and university settings, and Coder 3 relied on what he called "technological reflexivity," using his knowledge of institutional processes and fairness to judge AI outputs. In all cases, when AI conflicts arose, human codes and interpretations were upheld. Table 5 summarises these shared perspectives and illustrates them with exact examples from each coder.

Table 5
*Similarities in Researcher Actions and Perspectives During AI-Code Refinement*

| Shared action or perspective | Coder 1's action or example | Coder 2's action or example | Coder 3's action or example |
|---|---|---|---|
| Initial acceptance of the AI structure | Described the AI Phase 1 summary as "an added approach… I love this," acknowledging its usefulness for structuring the data. | Reported that the AI "did an incredibly amazing job" in bringing down the time required for manual analytic tasks. | Described the AI's ability to associate words with particular meanings as "above average." |
| Identifying a lack of contextual depth | Critiqued AI codes for being "too long" and needing "much simpler ways" to maintain contextual meaning. | Observed that AI codes "lacked emotional nuance" and neutralised emotionally charged language. | Pointed out that the AI "picked the literal meanings and not the implicit meanings," missing the broader institutional context. |
| Using lived experience as an audit tool | Used her experience as a tutor in a low to middle-income country to judge whether codes reflected actual classroom realities. | Refined interpretations, such as "extra time", are based on his own teaching experiences in basic and university education contexts. | Applied "technological reflexivity" by drawing on personal experiences with institutional processes and fairness to reinterpret AI outputs. |
| Upholding human judgment over AI output | Stated that if the AI "did not maintain the contextual meaning, I deleted that code" and replaced it with a human code. | Affirmed that even with AI support, he "remained the final interpreter of meaning" in the analysis. | Resolved discrepancies by "upholding human codes over AI codes," especially where AI interpretations clashed with contextual understanding. |
| Refining theoretical or academic language | Shifted from AI's general "ethical concern" to "unprofessional," aligning language with everyday professional discourse. | Revised terms such as "epistemological disconnects" to "epistemological disjuncture" for more precise institutional accuracy. | Adjusted codes to reflect "institutional practices and fairness," aligning AI outputs with his critical stance on equity and justice. |

## 4.4. CODERS REFLECTIONS ON HUMAN–AI COLLABORATION

Analysis of coders' reflections reveals four interrelated themes that characterise how human analysts experienced and negotiated collaboration with AI during inductive thematic analysis. Across all themes, AI functioned as a procedural and organisational scaffold, while humans retained interpretive authority, particularly in relation to contextual meaning, emotional nuance, and theoretical positioning.

### 4.4.1. Structural Efficiency versus Contextual Depth

All three researchers recognised AI's strength in accelerating analytic work and structuring large qualitative datasets. However, they consistently noted limitations in the AI's capacity to capture emotional nuance and implicit meaning. Fr example, Coder 1 acknowledged the organisational value of AI but emphasised the need for human depth, noting that AI "organize[s] and surface[s] key themes," yet often "neutralize[s] emotionally charged language." Similarly, Coder 2 described the AI as doing an "incredibly amazing job" in reducing manual workload and drawing meaning from transcripts, while still recognising moments where interpretations required human refinement. Coder 3 was more critical, observing that the AI's "sense of contextual meaning is narrow as compared to that of humans" and that it tended to "pick the literal meanings and not the implicit meanings."

### 4.4.2. Interpretive Stance and Lived Experience as Analytic Filters

Researchers' epistemological orientations and professional experiences strongly shaped how they evaluated AI-generated codes. For instance, Coder 2 explicitly aligned with a constructivist stance, describing meaning as "co-constructed between the participants and the researcher," which led them to accept many AI codes as representative of participant experiences. Coder 1, working from a constructivist orientation and personal teaching experience, revised AI outputs to foreground themes such as improvisation and professional legitimacy that were initially treated neutrally by the AI. Coder 3 adopted a critical theorist position, actively searching for "underlying institutional dynamics" and "emancipatory goals." While they described being impressed when the AI acted as a "critical analyst," they rejected codes that failed to connect individual statements to broader "power dynamics."

### 4.4.3. Differential Reliability of Code Types (In-Vivo vs. Gerunds)

Across coders, in-vivo (verbatim) codes were consistently viewed as the most reliable analytic starting point. This was explicit as Coder 1 reported that in-vivo codes were preferable because they "preserved the participant's exact language," whereas gerund codes often required revision to recover "contextual meaning." Coder 2 similarly affirmed the value of verbatim coding but observed that some AI-generated codes "were just too long to be codes" and therefore required simplification. Coder 3 strongly favoured "exact keyword (verbatim) coding" for capturing meaningful data segments and expressed discomfort with gerund codes such as "reinforcing dissatisfaction due to misfit," which they felt could alternatively reflect an "unwillingness to adapt."

## 5. DISCUSSION

The findings contribute to ongoing debates in thematic analysis and AI-assisted qualitative research by empirically examining how human analysts and generative AI interact during inductive thematic analysis (De Paoli, 2024; Lee et al., 2024). This aligns with prior human–computer interaction research emphasizing that GenAI is useful support but still requiring human interpretation, trust calibration, and iterative review (Wen et al., 2025; Yan et al., 2023). Building on prior work in reflexive thematic analysis (Braun & Clarke, 2006, 2021) and recent studies on GenAI-supported qualitative coding (Perkins & Roe, 2024; Lee et al., 2024), the findings clarify how and where epistemic authority is exercised when AI is integrated into analytic workflows. Consistent with prior studies, the findings show that AI is most effective as a procedural and organisational scaffold rather than as an independent analyst. Similar to Perkins and Roe (2024) and De Paoli (2024), the ITA tool supported rapid data familiarisation, initial code generation, and theme clustering. However, the present study extends this literature by demonstrating that procedural support alone does not constitute analytic interpretation. Instead, interpretive authority remained with human analysts, who treated AI outputs as provisional and revisable. This finding aligns with Braun and Clarke's (2006, 2021) position that thematic analysis is an iterative, judgment-driven practice rather than a mechanical extraction of meaning. While GenAI can enhance efficiency and consistency, it does not replace the interpretive labour required to construct themes that are theoretically meaningful and contextually grounded.

A central contribution of this study is the identification of modification, deletion, rejection, insertion, and commenting as systematic analytic actions through which researchers govern AI outputs. Prior research has acknowledged the need for human verification and refinement (Morgan, 2023; Lee et al., 2024) but has rarely specified how this oversight is enacted in practice. The present findings address this gap by making epistemic labour visible through interaction logs and revision traces. These actions operationalise what human-in-the-loop scholars have described as oversight and accountability (Amershi et al., 2014; Shneiderman, 2020). Rather than assuming that human judgment implicitly moderates AI outputs, this study shows that epistemic authority is exercised through observable, recurring analytic practices that protect trace-to-text integrity, contextual accuracy, and methodological rigor.

The finding that verbatim (in-vivo) codes were consistently viewed as the most reliable analytic foundation aligns with Naeem et al.'s (2023) emphasis on exact keyword extraction and traceability. While prior studies report that GenAI can infer basic themes (Morgan, 2023; De Paoli, 2024), this study demonstrates that abstraction beyond verbatim coding often introduces interpretive risk. Gerund-based and categorical codes frequently required human intervention to recover implicit meanings, emotional nuance, and institutional context. This supports critiques that GenAI performs well at surface-level pattern detection but struggles with deeper interpretive work (Lee et al., 2024). The findings therefore reinforce arguments that inductive thematic analysis must prioritise verbatim grounding before abstraction, particularly when AI systems are involved. This finding resonates with Nowell et al.'s (2017) concerns about subjectivity and transparency in thematic analysis, while also demonstrating how AI-assisted workflows can make such interpretive moves explicit and auditable rather than implicit.

The findings further align with Braun and Clarke's (2021) emphasis on reflexivity by showing that coders' epistemological orientations and professional histories shaped how AI outputs were evaluated. Unlike prior GenAI studies that treat researchers as interchangeable validators, this study demonstrates that analytic sensemaking is distributed across humans, tools, and representational artefacts, consistent with distributed cognition theory (Hutchins, 1995; Grinschgl & Neubauer, 2022). These findings support and extend existing literature by clarifying that GenAI enhances efficiency and structure but does not resolve longstanding challenges in thematic analysis related to interpretation, context, and epistemic judgment. Rather than offering a universal workflow, AI-assisted thematic analysis requires explicit frameworks, such as HACITA, that foreground human judgment, reflexivity, and analytic accountability.

## 6. LIMITATIONS AND FUTURE RESEARCH

This study has several limitations. To begin with, it involved a small number of experienced qualitative researchers and a limited set of interview transcripts, which constrains the generalisability of the findings to other researcher populations, disciplinary contexts, or levels of qualitative expertise. The study also focused on analytic processes rather than substantive analytic outcomes; it did not evaluate whether AI-assisted analysis produces higher-quality themes compared to fully human analysis. In addition, the analysis examined interaction with a single AI-assisted analytic tool configured around inductive thematic analysis. Different models, prompting strategies, or interface designs may produce different forms of human–AI collaboration. Finally, although the transcripts were situated within a Ghanaian education context, all coders shared familiarity with this setting, which may have shaped the types of revisions and judgments observed.

Future research can extend this work in several ways. Studies comparing novice and expert researchers would help clarify how analytic actions such as rejection, insertion, or reframing develop with methodological expertise. Comparative research examining analytic outcomes across human-only, AI-assisted, and hybrid workflows would complement the present process-oriented focus. Further work is also needed to examine design features and interface affordances, such as reflexive prompts and audit trails, that support epistemic control and transparency. Finally, extending this analytic framework to other qualitative methods and cultural contexts would help determine the broader applicability of human–AI collaborative analytic practices.

## 7. CONCLUSION

The present study examined human–AI collaboration during inductive thematic analysis, shifting attention away from AI outputs as end products and toward analytic interaction as an empirical object of study. Grounded in reflexive thematic analysis using HACITA, human-in-the-loop AI, and distributed cognition, the study adopted a qualitative, process-oriented design that treated interaction logs, revision traces, and researcher reflections as primary data. Rather than evaluating whether AI-generated themes were "correct," the analysis focused on how researchers engaged with AI suggestions, how they evaluated and revised them, and how analytic reasoning unfolded across humans, prompts, and artifacts. The findings show that AI functioned as a procedural and organisational scaffold, supporting familiarisation, verbatim coding, and theme clustering, but did not operate as an autonomous analyst. Across all analytic phases, human researchers consistently retained epistemic authority, enacting judgment through five recurring analytic actions: modification, deletion, rejection, insertion, and commenting. These actions constituted systematic analytic labour through which contextual meaning, emotional nuance, institutional relevance, and trace-to-text integrity were preserved. Verbatim (in-vivo) coding emerged as the most reliable foundation for human-AI collaboration, while gerund-based and categorical codes frequently required human refinement to recover implicit meaning and lived experience. Researchers repeatedly corrected AI literalism, reframed neutral language to reflect professional strain and institutional realities, and realigned codes with contextual knowledge of Ghanaian education.

## 8. REFERENCES

Acheampong, K. O., & Nyaaba, M. (2024). Review of qualitative research in the era of generative artificial intelligence. *SSRN Electronic Journal*. https://doi.org/10.2139/ssrn.4686920

Akbar, A. M., Khan, A. A., & Liang, P. (2023). *Ethical aspects of ChatGPT in software engineering research*. arXiv. https://doi.org/10.48550/arXiv.2306.07557

Andersen, J. P., Degn, L., Fishberg, R., Graversen, E. K., Horbach, S. P., Schmidt, E. K., ... & Sørensen, M. P. (2025). Generative artificial intelligence (GenAI) in the research process– A survey of researchers' practices and perceptions. *Technology in Society, 81*, Article 102813.

Braun, V., & Clarke, V. (2021). *Thematic analysis: A practical guide*. SAGE Publications.

Braun, V., & Clarke, V. (2024). Thematic analysis. In *Encyclopedia of quality of life and well-being research* (pp. 7187–7193). Springer Nature Switzerland.

Bryda, G., & Costa, A. P. (2024). Transformative technologies: Artificial intelligence and large language models in qualitative research. *Revista Baiana de Enfermagem, 38*. https://doi.org/10.18471/rbe.v38.61024

Burleigh, C., & Wilson, A. M. (2024). Generative AI: Is authentic qualitative research data collection possible? *Journal of Educational Technology Systems, 53*(2), 89–115. https://doi.org/10.1177/00472395241270278

Cárdenas, S., & Valcárcel, M. (2005). Analytical features in qualitative analysis. *TrAC Trends in Analytical Chemistry, 24*(6), 477–487. https://doi.org/10.1016/j.trac.2005.03.006

Chen, M., Li, Y., & O'Rourke, E. (2024). Understanding the reasoning behind students' self-assessments of ability in introductory computer science courses. In *Proceedings of the 2024 ACM Conference on International Computing Education Research* (pp. 1–13). https://doi.org/10.1145/3632620.3671094

Cheung, K. K. C., & Tai, K. W. H. (2021). The use of intercoder reliability in qualitative interview data analysis in science education. *Research in Science & Technological Education, 41*(3), 1155–1175. https://doi.org/10.1080/02635143.2021.1993179

Clarke, V., & Braun, V. (2016). Thematic analysis. *The Journal of Positive Psychology, 12*(3), 297–298. https://doi.org/10.1080/17439760.2016.1262613

Cohen, J. (1960). A coefficient of agreement for nominal scales. *Educational and Psychological Measurement, 20*(1), 37–46.

Dahal, N. (2024). How can generative AI (GenAI) enhance or hinder qualitative studies? A critical appraisal from South Asia, Nepal. *The Qualitative Report, 29*(3), 722–733.

De Paoli, S. (2024). Performing an inductive thematic analysis of semi-structured interviews with a large language model: An exploration and provocation on the limits of the approach. *Social Science Computer Review, 42*(4), 997–1019.

Draper, N. R., & Smith, H. (1998). *Applied regression analysis* (3rd ed., Vol. 326). John Wiley & Sons.

Finlay, L. (2021). Thematic analysis: The 'good,' the 'bad,' and the 'ugly.' *European Journal for Qualitative Research in Psychotherapy, 11*, 103–116. https://ejqrp.org/index.php/ejqrp/article/view/136

Gao, J., Guo, Y., Lim, G., Zhan, T., Zhang, Z., Li, T. J. J., & Perrault, S. T. (2023). *CollabCoder: A GPT-powered workflow for collaborative qualitative analysis*. arXiv. https://doi.org/10.48550/arXiv.2304.07366

Liu, X., Pankiewicz, M., Gupta, T., Huang, Z., & Baker, R. S. (2024). *A step towards adaptive online learning: Exploring the role of GPT as virtual teaching assistants in online education* [Manuscript submitted for publication].

Liu, X., Zhang, J., Barany, A., Pankiewicz, M., & Baker, R. S. (2024). Assessing the potential and limits of large language models in qualitative coding. In Y. J. Kim & Z. Swiecki (Eds.), *Advances in Quantitative Ethnography. ICQE 2024. Communications in Computer and Information Science* (Vol. 2278). Springer, Cham. https://doi.org/10.1007/978-3-031-76335-9_7

Morgan, D. L. (2023). Exploring the use of artificial intelligence for qualitative data analysis: The case of ChatGPT. *International Journal of Qualitative Methods, 22*. https://doi.org/10.1177/16094069231211248

Naeem, M., Ozuem, W., & Howell, K. (2023). Understanding 'dark social' from a digital marketing perspective: A novel six-step approach for conceptual model development in qualitative research. *Qualitative Market Research: An International Journal, 26*(3), 297–316.

Nowell, L. S., Norris, J. M., White, D. E., & Moules, N. J. (2017). Thematic analysis: Striving to meet the trustworthiness criteria. *International Journal of Qualitative Methods, 16*(1). https://doi.org/10.1177/1609406917733847

Nyaaba, M., Kyeremeh, P., Majialuwe, E. K., Owusu-fordjour, C., et al. (2024). Generative AI in academic research: A descriptive study on awareness, gender usage, and views among pre-service teachers. *Journal of AI, 8*(1), 45–60. https://doi.org/10.61969/jai.1400867

Perkins, M., & Roe, J. (2024). Academic publisher guidelines on AI usage: A ChatGPT-supported thematic analysis. *F1000Research, 12*, 1398. https://doi.org/10.12688/f1000research.142411.2

Petre, G.-E., & Costa, A. P. (2024). Advancing qualitative research: Insights from the 7th World Conference on Qualitative Research. *Social Sciences, 13*(1), 68. https://doi.org/10.3390/socsci13010068

Prescott, M. R., Yeager, S., Ham, L., Rivera Saldana, C. D., Serrano, V., Narez, J., ... & Montoya, J. (2024). Comparing the efficacy and efficiency of human and generative AI: Qualitative thematic analyses. *JMIR AI, 3*, e54482.

Qiao, S., Fang, X., Garrett, C., Zhang, R., Li, X., & Kang, Y. (2024). Generative AI for qualitative analysis in a maternal health study: Coding in-depth interviews using large language models (LLMs). *medRxiv*. https://doi.org/10.1101/2024.09.24.24314291

Rahman, M. M., & Watanobe, Y. (2023). ChatGPT for education and research: Opportunities, threats, and strategies. *Applied Sciences, 13*(9), 5783. https://doi.org/10.3390/app13095783

Sallam, M. (2023). ChatGPT utility in healthcare education, research, and practice: Systematic review on the promising perspectives and valid concerns. *Healthcare, 11*(6), 887. https://doi.org/10.3390/healthcare11060887

Wohlrab, R., Knauss, E., Steghöfer, J. P., et al. (2020). Collaborative traceability management: A multiple case study from the perspectives of organization, process, and culture. *Requirements Engineering, 25*, 21–45. https://doi.org/10.1007/s00766-018-0306-1

Xiao, Z., Yuan, X., Liao, Q. V., Abdelghani, R., & Oudeyer, P. Y. (2023). Supporting qualitative analysis with large language models: Combining codebook with GPT-3 for deductive coding. In *Companion Proceedings of the 28th International Conference on Intelligent User Interfaces* (pp. 75–78). Association for Computing Machinery.

Zambrano, A. F., Liu, X., Barany, A., Baker, R. S., Kim, J., & Nasiar, N. (2023). From nCoder to ChatGPT: From automated coding to refining human coding. In G. Arastoopour Irgens & S. Knight (Eds.), *Advances in Quantitative Ethnography. ICQE 2023. Communications in Computer and Information Science* (Vol. 1895). Springer, Cham. https://doi.org/10.1007/978-3-031-47014-1_32

*Under Review* JDEM# 9. APPENDICES

## 9.1. APPENDIX A

*Categorized Researcher Actions and Revisions to AI-Generated Codes*

| Researcher | Action category | Original AI content/context | Action taken (revision or insertion) | Researcher comment/rationale |
|---|---|---|---|---|
| Coder 1 | Commenting and approval | Phase 1 summary | Maintained original summary | Described the AI summary as "an added approach to the entire coding… I love this," acknowledging its structural value. |
| | Suggested insertion and simplification | "The lecturer insisted we buy his handout…" | Suggested gerund: "Buy the lecturer's handout or fail the course." | Noted that the AI code was too long, but helpful for arriving at the correct gerund. |
| | Substitution of terminology | AI rationale: "…signals ethical concern." | Revised the key term to "unprofessional." | Agreed that the practice raised ethical issues, but considered "unprofessional" a more precise descriptor. |
| | Deletion and replacement of codes | AI-generated codes across several segments | Deleted AI codes and replaced them with human-generated codes. | Reported that she removed AI codes whenever they failed to maintain contextual meaning and used human codes instead. |
| Coder 2 | Nuanced edit: emotional and professional | "That takes extra time and effort." | Revised to "emotional labor and professional strain." | Sought to capture the psychological impact and professional burden on underprepared educators. |
| | Nuanced edit: adaptive professionalism | "I used the knowledge and experience I had…" | Revised to "improvised teaching practices based on lived experience… adaptive professionalism." | Recast a literal description as a form of professional adaptability. |
| | Structural edit | "indicates a structural issue in the education sector." | Revised to "identifies a structural bottleneck in teacher education." | Used more specific academic terminology to describe systemic failure in teacher preparation. |
| | Nuanced edit: | "The students… challenge me to | Revised to "signals reciprocal learning." | Highlighted that student curiosity and questioning |
| | mutual learning | prepare thoroughly." | | drive teacher rigour and learning. |
| | Constraint on professional development | "It is easy to remain stagnant there." | Reframed as a condition that "constrains professional development." | Emphasised that the system limits opportunities for growth, not just individual choices. |
| | Epistemic edit | "Highlights epistemological disconnects." | Revised to "identifies an epistemological disjuncture." | Refined the language to reflect hierarchical institutional gaps at the epistemological level. |
| | New code insertion | N/A | Inserted code such as "recognizing systemic gaps in teacher preparation." | Added a new interpretive category when AI did not fully capture systemic concerns. |
| | Data insertion for audit trail | N/A | Inserted multiple verbatim participant quotes into rationale columns. | Strengthened the audit trail and grounded interpretations in participants' own words. |
| Coder 3 | Categorical deletion and replacement | Category label "Professional (dis)engagement." | Deleted and replaced with "Professional challenges." | Argued that "disengagement" implies withdrawal from duties, which did not reflect the participant's continued work. |
| | Refining insertion: professional intent | "Valuing comfort in one's specialization." | Revised to "insisting on preferred area of specialization." | Emphasised the participant's active demand for professional alignment, not mere comfort. |
| | Refining insertion: systemic outcomes | "Observing systemic misallocation." | Revised to "ensuing outcomes of the current allocation process." | Clarified the link between existing allocation processes and resulting problems. |
| | Nuanced edit: workload and strain | "Undertaking extra preparatory labour." | Revised to "additional workload and professional strain." | Broadened the focus from extra work alone to include mental and professional stress. |
| | Simplification of category label | "Suggesting practical, hierarchical reform." | Shortened to "suggesting practical reform." | Reduced complexity to promote clearer and more accessible coding. |
| | Clarifying note | Code located within the mathematics education segment | Added note: "Since the issue goes beyond just mathematics." | Ensured that the interpretation was not restricted to a single subject area. |

| Final judgement on AI versus human codes | Conflicts between AI and human codes across several instances | Upheld human codes over AI codes. | Applied a consistent policy of prioritising human contextual understanding over AI literalism. |